\documentclass[10pt, a4paper]{article}

\usepackage[final]{lrec2026} 

\usepackage{booktabs}
\usepackage{threeparttable}
\usepackage{adjustbox}
\usepackage{authblk}
\usepackage[toc,page]{appendix}

\title{PsihoRo: Depression and Anxiety Romanian Text Corpus}

\name{Alexandra Ciobotaru$^{1,2}$, Ana-Maria Bucur$^{3}$, Liviu P. Dinu$^{1,4}$} 

\address{$^{1}$University of Bucharest, $^{2}$Druid AI, $^{3}$Università della Svizzera italiana, \\
$^{4}$Human Language Technologies Research Center \\
         $^{1,4}$90 Panduri Street, Bucharest, Romania \\
         $^{2}$1-5 Costache Negri Street, Bucharest, Romania, \\
         $^{3}$Via Buffi 13, Lugano, Switzerland \\
         alexandra.zbaganu@s.unibuc.ro, ana-maria.bucur-cosma@usi.ch, ldinu@fmi.unibuc.ro\\}

\abstract{Psychological corpora in NLP are collections of texts used to analyze human psychology, emotions, and mental health. These texts allow researchers to study psychological constructs, identify patterns related to mental health problems and analyze emotional language. However, collecting accurate mental health data from social media can be challenging due to the assumptions made by data collectors. A more effective approach involves gathering data through open-ended questions and then assessing participants' mental health status using self-report screening surveys. This method was successfully employed for English, a language with a lot of psychological NLP resources. However, the same cannot be stated for Romanian, which currently has no open-source mental health corpus. To address this gap, we have collected the first open-source corpus focused on depression and anxiety in Romanian, by utilizing a form with 6 open-ended questions along with the standardized PHQ-9 and GAD-7 screening questionnaires. Although the PsihoRo corpus contains texts from only 205 respondents, it represents an important first step toward understanding and analyzing mental health issues within the Romanian population. We employ statistical analysis, text analysis using Romanian LIWC, emotion detection, and topic modeling to identify the most important features of this newly introduced resource for the NLP community. The data is publicly available at \url{https://huggingface.co/datasets/Alegzandra/PsihoRo}.
\\ \newline 
}

\begin{document}

\maketitleabstract

\section{Introduction}


Mental health research has increasingly used data from social media platforms, which often present a curated view of individuals' personal lives. While social media can provide valuable insights for mental health studies \cite{liu2022head,rai2024key,Bucur2024state}, there is a risk of positivity bias, where users tend to share mostly positive aspects of their experiences \cite{salopek2024toxic}. Another important source for mental health research is clinical data, which typically contains sensitive information and is often not publicly accessible \cite{puspo-etal-2026-mental}.

Most research on mental health is conducted using data in English, which may overlook important nuances present in other languages. Previous studies suggest that features predictive of mental health issues in English may not translate effectively to other languages or cultures \cite{rai2024key,bucur2025survey}. For example, the use of the first-person pronoun ``I'' has been identified as a strong predictor of depression \cite{rude2004language}, and this feature is frequently analyzed in English-based research. However, the relationship between pronoun usage and depression severity can vary significantly across different demographic groups and languages \cite{rai2024key}. While the pronoun ``I'' is often used as an indicator of depression in English, its use in other languages requires special consideration due to linguistic differences. For instance, previous research has shown that this association has not been observed among Romanian speakers \cite{trifu2024linguistic}. Romanian is a pro-drop language, which allows for ambiguity in subject pronoun usage and may contribute to a lower frequency of the personal pronoun ``I'' in everyday communication.

Despite the availability of data in the Romanian language related to emotions, such as the RED dataset \cite{ciobotaru2021red}, REDv2 \cite{ciobotaru-etal-2022-red}, and the BRIGHTER multilingual dataset \cite{muhammad-etal-2025-brighter}, which includes Romanian data, there is a scarcity of mental health-related resources. Very few clinical data sources \cite{trifu2024linguistic} and posts from medical forums \cite{briciu2018studying,duduau2022development} exist, but unfortunately are not publicly accessible.

This study aims to address this gap by creating the first publicly available dataset in Romanian, specifically designed to analyze emotional expressions in the context of depression and anxiety. By analyzing the linguistic characteristics and emotional patterns in Romanian, our research aims to contribute to a deeper understanding of mental health expressions in languages other than English. Through this initiative, we seek to promote more inclusive research in mental health, paving the way for targeted interventions and support across diverse linguistic communities.
\section{Related Work}

Previous research has gathered open-ended responses from participants to explore how expressions of mental health influence language. \citet{Yoon2024Predicting} created open-ended personality questions to explore how natural language processing can aid in predicting neuroticism from people’s responses. The authors gathered over 30,000 sentences written by 425 adults in Korean and used KoBERT to see which types of questions best revealed neuroticism and related traits like depression and dependency. \citet{vanderVegt2023} created the multi-modal Real World Worry Waves Dataset (RW3D), in English, which tracks the psychological effects of the COVID-19 pandemic in the UK over three consecutive years, 2020, 2021 and 2022, through repeated surveys containing also free-text responses.  \citet{Dai2022Explainable} used open-ended interview responses from more than 58,000 job candidates along with self-assessed personality (HEXACO) and employed “InterviewBERT” to deduce personality traits from textual in English. \citet{Sikstrom2023Assessment} used open-ended questionnaires to study young and middle-aged adults’ responses regarding their own mental health, by employing NLP computational methods. The authors analyzed 1,106 responses and found semantic differences between the two study groups.

In terms of combining emotion detection and LIWC term analysis in mental health texts, several studies employ such analysis in English: \citet{Rahman2024_DepressionEmo}, \citet{ZHANG2023231}, \citet{Alshouha2024_BioEmoDetector}. \citet{BEECH2025113} combined LIWC features \cite{boyd2022development}, sentiment analysis and NLP to explore the relationship between language and mental health issues such as depression, anxiety, and stress during the COVID-19 pandemic. They analyzed written narratives from 883 participants, alongside their DASS-21 scores, and found an association between increased symptom severity and greater use of first-person pronouns and negative emotion words. \citet{Kjell2021AItextDepression} collected responses in English from 411 participants who answered word-response questions related to their mental health and completed the GAD-7 and PHQ-9 rating scales. Their analyses show that individuals’ language reflects their mental health status, and combining verbal responses with quantitative measures can enhance the comprehension and assessment of the risk of depressive and anxiety symptoms in both clinical practice and research contexts.

While several datasets have been created using text gathered from open-ended questionnaires, combined with screening tools to assess the risk of mental health issues, most of these resources are focused on the English language. Currently, there are no open-source psychological corpora available in Romanian, and this was the main driving force behind our study, along with our interest in applying emotional analysis to such a corpus.

\section{Data Collection}

As opposed to social media collected data, which is collected without participants' consent, we collected data from participants who provided their informed consent for data collection \cite{benton2017ethical}. The data collection methodology consists of a completely anonymous survey, organized in three parts. The first part of the survey consists of six open-ended questions, designed to gather extensive written responses. Out of these six questions, three touch on positive topics and three on negative topics. The survey was written in Romanian for Romanian speakers only. We further present the six open-ended questions, in both English and Romanian:

\begin{enumerate}
    \item To what extent do you think social media affects your emotional state, especially when you see negative news? ``În ce măsură crezi că rețelele sociale îți afectează starea emoțională, mai ales când vezi știri negative?'' (negative)
    \item What is an important lesson about resilience and adaptation that you have learned recently? ``Care este o lecție importantă despre reziliență și adaptare pe care ai învățat-o recent?'' (positive)
    \item How much do you think politics affects people's mental health? Can you describe a time when you felt stressed because of political decisions? ``Cât de mult crezi că politica afectează sănătatea mentală a oamenilor? Poți să detaliezi un moment în care ai simțit stres din cauza deciziilor politice?'' (negative)
    \item What role do you think community or social support plays in maintaining mental health during difficult times? ``Ce rol crezi că are comunitatea sau sprijinul social în menținerea sănătății mentale în vremuri dificile?'' (positive)
    \item How were you affected by discussions about war, the economic crisis, elections, or other tense topics? ``Cum ai fost afectat de discuțiile despre război, criza economică, alegeri sau alte subiecte tensionate?'' (negative)
    \item How do you recharge your batteries after a stressful period? ``Cum îți reîncarci bateriile după o perioadă stresantă?'' (positive)
\end{enumerate}

Usually, psychological corpora for NLP are gathered from social media through searching for different keywords in online posts that could indicate depression and anxiety. However, after consulting with medical professionals in the psychiatric field, we have concluded that this method of collecting psychological data may not be entirely accurate. Thus, we added two more parts to our survey: one assessing respondents' risk of depression using the PHQ-9 \cite{kroenke2001phq9}, and another assessing their risk of anxiety using the GAD-7 \cite{spitzer2006gad7}. Both GAD-7 and PHQ-9 contain questions (7 for GAD-7 and 9 for PHQ-9) that ask the participants how often they have experienced specific problems, and each item is answered on a 4-point Likert scale: Not at all = 0, Several days = 1, More than half the days = 2, Nearly every day = 3. The severity cut-offs for depression based on PHQ-9 are: 0-4: Minimal, 5-9: Mild, 10-14: Moderate, 15-19: Moderately severe, 20-27: Severe, while the severity cut-offs for anxiety based on GAD-7 are: 0-4: Minimal, 5-9: Mild, 10-14: Moderate, 15-19: Moderately severe, 20-27: Severe.

The resulting questionnaire is completely anonymous, and no personal data is requested or stored (such as name, surname, email address, or IP). We did not collect demographic data for the same reason: to build trust with respondents and keep the time required to complete the form short. 

The resulting data consists of 205 responses, collected over an eight-month period from March 2025 to October 2025. This time frame includes two major stressful events for Romanians--presidential elections and the tax increase burden generated by the country's new leadership.

\section{Methodology}
\label{sec:methodology}
Our research consists of gathering insights from PsihoRo using descriptive statistics to analyze the correlation between depression and anxiety levels among respondents across three main population groups, computed based on severity cut-offs. According to the literature in psychology \cite{kroenke2001phq9}, a PHQ-9 score of 10 or higher indicates the potential presence of major depressive disorder and prompts further assessment. For the GAD-7 questionnaire, a cut-off score of 10 has also been proposed as a threshold for identifying potential generalized anxiety disorder, warranting further evaluation \cite{spitzer2006gad7}.

Based on these cut-off scores, three population groups were derived from our dataset:

\begin{itemize}
    \item Depression Risk Group: 37 respondents scored higher than 10 on PHQ-9 (18\%);
    \item Anxiety Risk Group: 36 respondents scored higher than 10 on GAD-7 (17.56\%);
    \item No Risk Group: 28 respondents scored 0 for at least one out of the two questionnaires (13.66\%).
\end{itemize}

Linguistic Inquiry and Word Count (LIWC) \cite{boyd2022development} is a lexicon and a text analysis software that provides psychologically meaningful categories and is widely used in computational mental health research \cite{rude2004language,liu2022head,hur2024language,trifu2024linguistic}. The official LIWC software is available for analyzing textual data in the English language; however, validated translations of the lexicon are also available in several languages. In our experiments, we use the Romanian version of LIWC (Ro-LIWC2015) \cite{duduau2022development}. The 2015 version of LIWC includes 98 categories that relate to linguistic dimensions and grammar, as well as psychological processes, including affective, cognitive, social, perceptual, and biological processes.

Previous research on the English language has indicated that individuals with mental health issues tend to exhibit specific language patterns. These patterns include a higher frequency of using the personal pronoun ``I'' \cite{rude2004language}, increased use of words with negative connotations \cite{fekete2002internet}, and a tendency to use words with absolute meanings \cite{al2018absolute}. These findings highlight the importance of language analysis for understanding mental health and motivate us to use LIWC to examine participants' language patterns. After checking for differences in LIWC categories across PHQ-9 and GAD-7 scores, we wanted to test their discriminatory power for classification. To this end, we employed a LightGBM classifier \cite{ke2017lightgbm} and trained it on LIWC results to predict a binary label for depression and anxiety, using a cut-off score of 10. The model achieved accuracies of 0.85 for depression and 0.83 for anxiety. To determine which LIWC features have the greatest influence on predictions, we used SHapley Additive exPlanations (SHAP) \cite{lundberg2017unified}.

To complement these analyses, we have created an emotion detection model by fine-tuning the Romanian BERT \cite{dumitrescu-etal-2020-birth} on the REDv2 dataset \cite{ciobotaru-etal-2022-red}, for multi-label text classification, using the Huggingface transformers pipeline.\footnote{\href{https://huggingface.co/docs/transformers}{https://huggingface.co/docs/transformers}} We used this model to predict the emotions in PsihoRo, thus gathering emotional information for the text written by each population group mentioned above. The resulting model achieved an F1 score of 66.85\% on the REDv2 test set.

We also performed topic modeling to identify the main themes in the PsihoRo corpus, using the \texttt{stm} R package. First, the texts were tokenized, lowercased, and cleaned by removing punctuation, numbers, symbols and stopwords.
Next, we created a Document-Feature Matrix (DFM) to represent term frequencies across texts. To enhance model performance and reduce noise, we excluded rare terms that appeared in fewer than 3 documents and the very common terms that appeared in more than 90\% of the texts. Further, the cleaned DFM was transformed into the format required by the Structural Topic Model (STM), and we trained the model with $K$ = 5 topics using the Spectral initialization method.
\section{Experiments and Results}






\subsection{PsihoRo Descriptive Statistics}

In Table \ref{stats}, we show mean, median (50\% percentile), min, max, 25\% and 75\% percentiles, and the standard deviation for both anxiety and depression scores. In the collected data, the maximum value for GAD-7 (anxiety) is 21, which is the maximum that can be achieved, and the maximum score for PHQ-9 (depression) is 23, out of the maximum possible of 27.

\begin{table}[t]
\centering
\label{stats}
\begin{tabular}{lrr}
\toprule
 & \textbf{PHQ-9} & \textbf{GAD-7} \\
\midrule
Count & 205 & 205 \\
Mean & 6.00 & 5.78 \\
SD & 4.83 & 4.95 \\
Min & 0 & 0 \\
25\% & 3 & 2 \\
50\% (Median)& 5 & 5 \\
75\% & 8 & 7 \\
Max & 23 & 21 \\
\bottomrule
\end{tabular}
\caption{\label{stats} Descriptive statistics for PsihoRo}
\end{table}

We analyzed the length of responses to determine whether individuals at risk of depression or anxiety tend to write less. Our findings indicate that this is not the case for Romanian. The computed T-statistic was 1.55 when comparing the depression risk group with the group not at risk, and 2.26 when comparing the anxiety risk group with the group not at risk. These results suggest that there is no significant difference in response length between these populations. 

\begin{figure}[t!]
    \centering
    \includegraphics[width=0.9\linewidth]{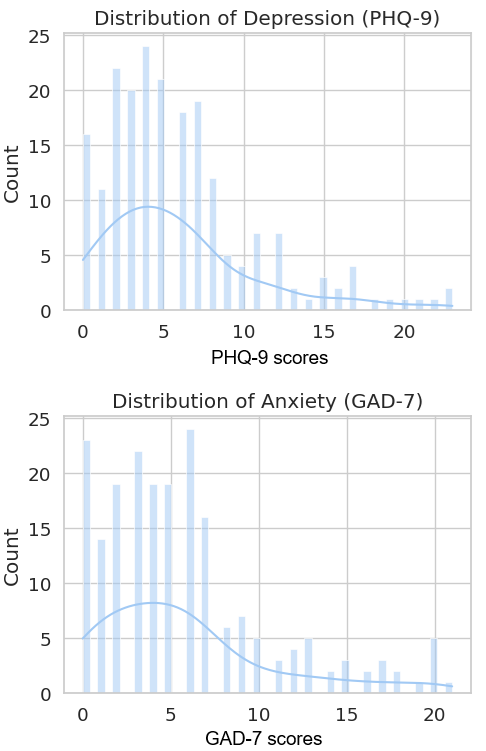}
    \caption{Distributions for GAD-7 and PHQ-9}
    \label{fig:distributions}
\end{figure}

In Figure \ref{fig:distributions}, we present the distribution plots of both PHQ-9 and GAD-7 scores. The data indicate that the distributions are skewed to the left, with most observations clustered at lower values and a longer tail extending toward the higher end of the scale. This suggests that the majority of respondents experience minimal to mild risk of depression and anxiety.

\begin{figure}
    \centering
    \includegraphics[width=0.8\linewidth]{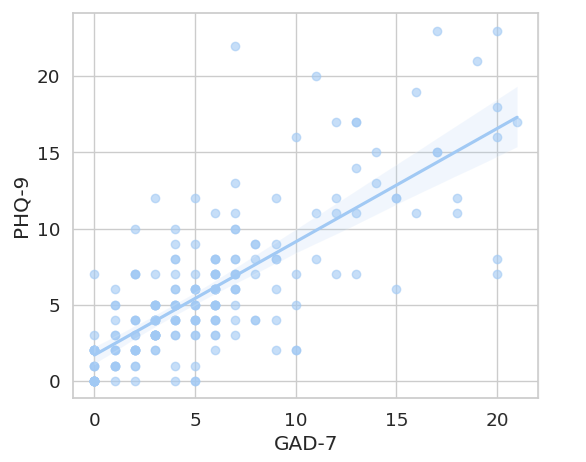}
    \caption{Correlation between Anxiety and Depression Scores.}
    \label{fig:correlation}
\end{figure}

The Pearson correlation coefficient ($r$) between the PHQ-9 and GAD-7 total scores is 0.761 ($p$<0.001), indicating that anxiety and depression are strongly correlated in our data. This finding is illustrated in Figure \ref{fig:correlation} and aligns with the results of several other studies \cite{doi:10.1176/appi.ajp.2020.20030305,Hoying2020,Jacobson2014}.


\begin{table}[t!]
\adjustbox{width=\linewidth}{%
\centering
\begin{tabular}{llrr}
\toprule
\textbf{LIWC Dimension}  &\textbf{Examples}& \textbf{PHQ-9} & \textbf{GAD-7} \\
\midrule
I  &I, me, mine& .018& -.046\\
We &we, us, our& .013& -.039\\
You &you, your& .006& .084\\
 Verb& eat, come, carry& -.087&-.091\\
 Past focus& ago, did, talked& -.064&-.075\\
 Present focus& today, is, now& -.019&-.034\\
 Future focus& may, will, soon& -.027&-.055\\
 Adjectives& free, happy, long& .118&.120\\
 Negations& no, not, never& -.156*&-.167*\\
 Positive emotion&love, nice, sweet& -.122&-.057\\
 Negative emotion&hurt, ugly, nasty& .028&.028\\
 Anxiety&worried, fearful& -.042&.010\\
 Anger&hate, kill, annoyed& .107&.105\\
 Sadness&crying, grief, sad& .074&.045\\
 Family&daughter, dad, aunt& .079&.145*\\
 Friends&buddy, neighbor& -.026&.043\\
 Female references&girl, her, mom& .061&.074\\
 Male references&boy, his, dad& .156*&.177*\\
 Comparatives&greater, best, after& .167*&.081\\
 Causation&because, effect& -.186**&-.180**\\
 Tentative &maybe, perhaps& .204**&.213**\\
 Insight& think, know& -.075&-.101\\
 Assent& agree, OK, yes& .009&.065\\
 Perceptual processes& look, heard, feeling& .009&.033\\
 Body&cheek, hands, spit& .200**&.180**\\
 Health& clinic, flu, pill& -.051&.035\\
 Achievement&win, success, better& -.163*&-.192**\\
 Leisure&cook, chat, movie& -.145*&-.169*\\
 Death&bury, coffin, kill& .138*&.029\\
\bottomrule
\end{tabular}}
\small
\textit{Note.} * $p < .05$, ** $p < .01$.

\caption{Correlations between PHQ-9, GAD-7, and LIWC dimensions.}
\label{tab:liwc_corr_tab}
\vspace{-5mm}
\end{table}

\begin{figure*}[h!]
    \centering
    \includegraphics[width=1\linewidth]{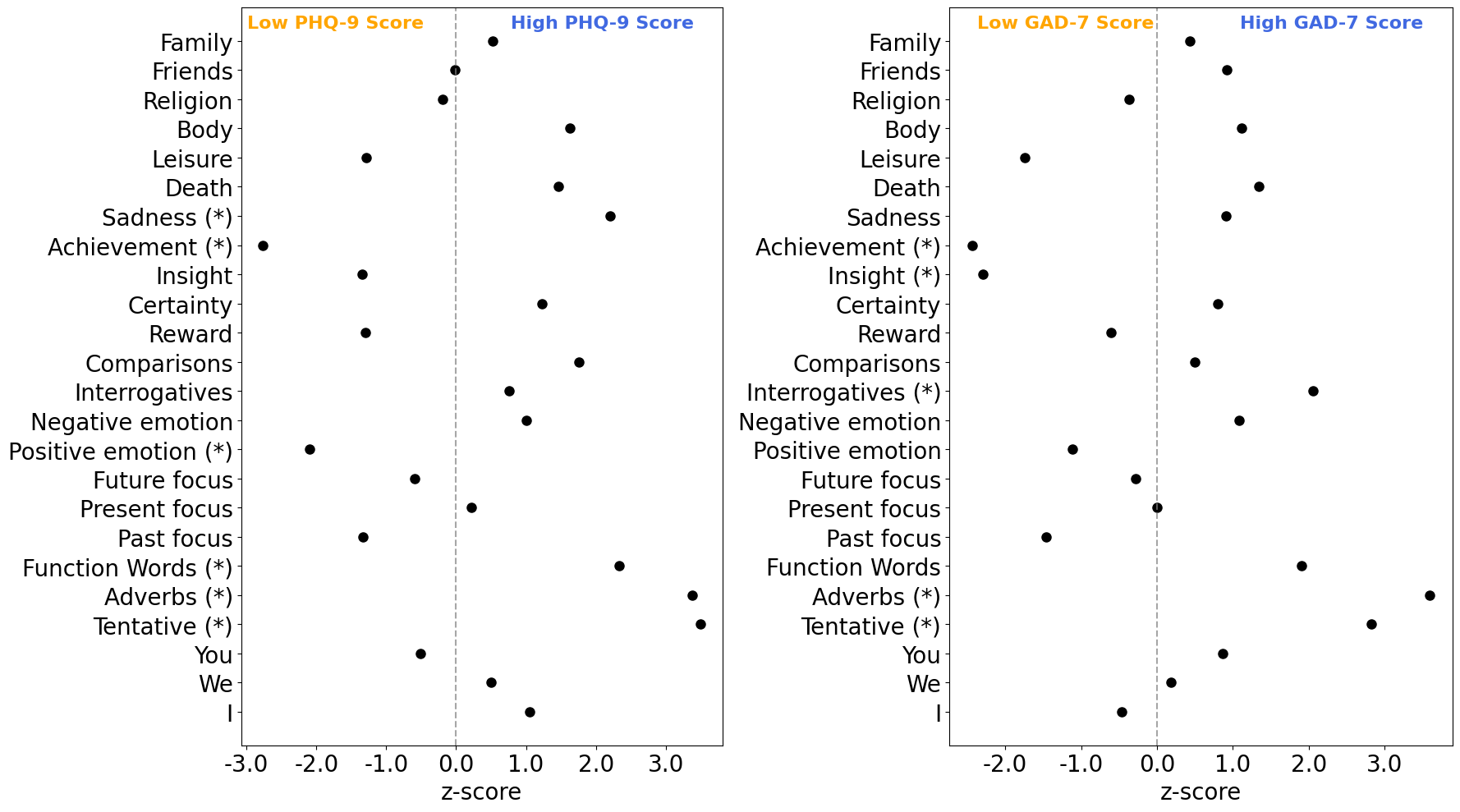}
    \caption{Differences in selected LIWC linguistic categories between participants with low vs. high PHQ-9 (left) and GAD-7 (right) scores. * denotes statistically significant differences ($p<0.05$) measured using the Mann-Whitney U test. Positive values indicate greater relative use among participants with higher symptom scores.}
    \label{fig:liwc}
\end{figure*}

\subsection{LIWC}
We employ LIWC to analyze language patterns in participants' open-ended responses and correlate them with scores from the PHQ-9 and GAD-7. In Table \ref{tab:liwc_corr_tab}, we present the Pearson correlation coefficients between LIWC categories and the symptom scores. Positive values indicate that higher symptom scores are associated with greater use of that linguistic category, while negative values indicate the opposite.

\begin{figure*}[t!]
    \centering
    \includegraphics[width=\linewidth]{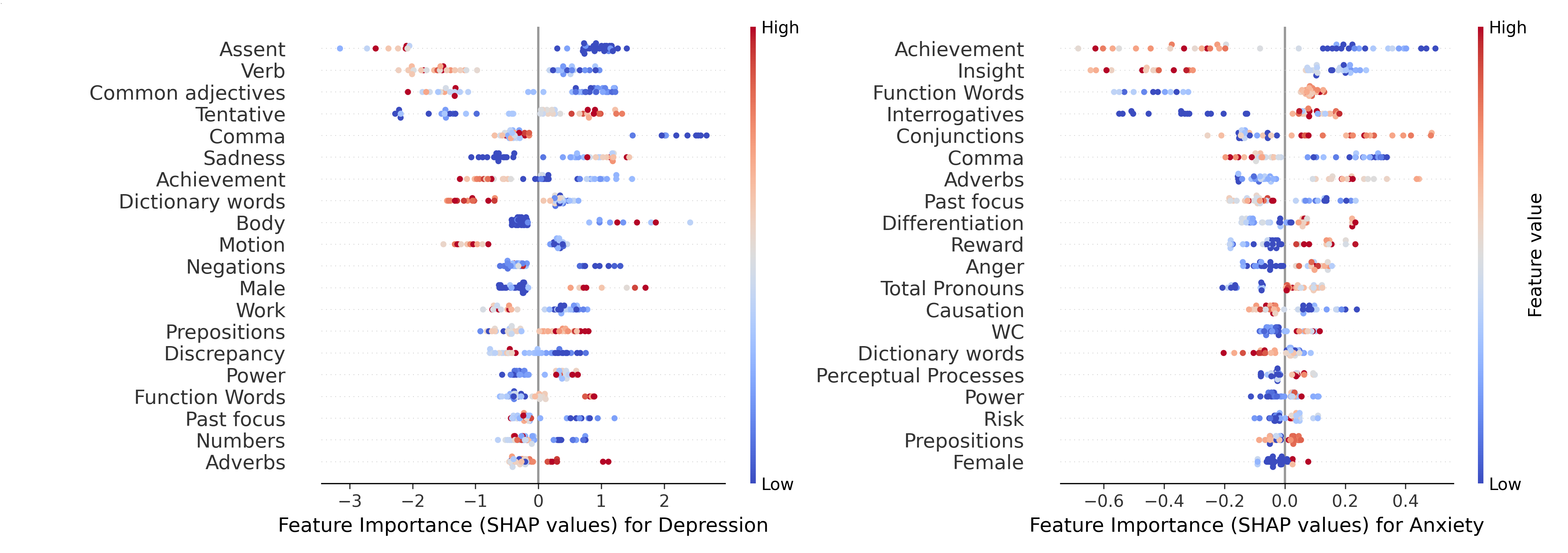}
    \caption{Feature importance for LIWC categories analyzed using SHAP.}
    \label{fig:shap}
\end{figure*}

Surprisingly, the use of negations is negatively correlated with both PHQ-9 and GAD-7 scores, suggesting that participants with higher depression and anxiety scores tend to use fewer negations in their responses. Individuals who score higher on the PHQ-9 and GAD-7 tend to use more words in the Tentative category, which indicates uncertainty, as evidenced by correlation coefficients of .204** for PHQ-9 and .213** for GAD-7. Comparatives also show a positive correlation with the depression score (.167*), indicating comparative thinking, which is often linked to depressive rumination \cite{nolen2008rethinking}. However, another marker for rumination, the high frequency of Past focus verbs, is not observed in our data. 
There is a positive correlation between the use of Body-related words and PHQ-9 (.200**) and GAD-7 (.180**) scores, indicating that participants with higher scores frequently use more language related to bodily sensations. This may reflect increased somatic awareness, a common trait in depression and anxiety \cite{penninx2013understanding}. Participants with higher PHQ-9 and GAD-7 scores discuss Achievement less frequently (-.163* and -.192**, respectively), which could indicate lower motivation. In addition, the negative correlation with Leisure-related words (-.145* and -.169*, respectively) may stem from anhedonia, a common symptom of depression \cite{grillo2016anhedonia}. Words in the Death category show a significant positive correlation with depression (.138*), but not with anxiety, indicating a greater preoccupation with mortality in participants with higher depression scores. In contrast, the Family category is positively correlated with anxiety scores (r = .145*), suggesting that individuals with higher anxiety are more focused on family-related issues.

To better understand the differences between higher and lower symptom scores for depression and anxiety, we conducted an additional analysis of the LIWC categories. According to the cut-off values for both the PHQ-9 and GAD-7, we define a low PHQ-9 score as less than 10, while a score of 10 or higher is considered high. In Figure \ref{fig:liwc}, the goal was to identify linguistic markers that differentiate individuals with higher vs. lower symptom scores. Statistically significant differences are indicated with an asterisk (*), as determined by the Mann-Whitney U test. Positive values indicate greater relative use among participants with higher symptom scores,
while negative values indicate greater use among low-symptom participants. The analysis produced results consistent with the correlation analysis, particularly regarding Achievement and tentativeness in relation to depression and anxiety. However, for depression, we observed a decreased use of Positive emotion words and an increased usage of Sadness words, which aligns with existing literature in psychology \cite{trifu2024linguistic}. In addition, the language of participants with high GAD-7 scores showed a higher frequency of words in the Interrogatives category, indicating a tendency to question or seek clarity.

To determine which LIWC features have the greatest influence on predictions, we used SHapley Additive exPlanations (SHAP) \cite{lundberg2017unified}, as explained in Section \ref{sec:methodology}, with the results presented in Figure \ref{fig:shap}. The importance of the top 20 features aligns with our correlation analysis and highlights the differences between low and high scores for both depression and anxiety. For predicting depression, low values in the LIWC features for Assent, Verb, and Common Adjectives, along with high values in Tentative words, contribute to the model's positive predictions. In the case of anxiety, low values in LIWC features related to Achievement and Insight, combined with high values in Function words, Interrogatives, and Conjunctions, also have a higher contribution to a positive prediction.

The LIWC analyses presented in this section demonstrate that lexical indicators can effectively capture underlying psychological patterns related to depression and anxiety. However, the most important feature used in mental health research, the pronoun ``I'', appears to be less useful for the Romanian language, likely because it is a pro-drop language. This finding is consistent with the work of \citet{trifu2024linguistic}, which also indicated that this marker does not generalize well in a sample of Romanian patients with major depression.

\subsection{Emotion Analysis}

\begin{figure*}[h!]
    \centering
    \includegraphics[width=1\linewidth]{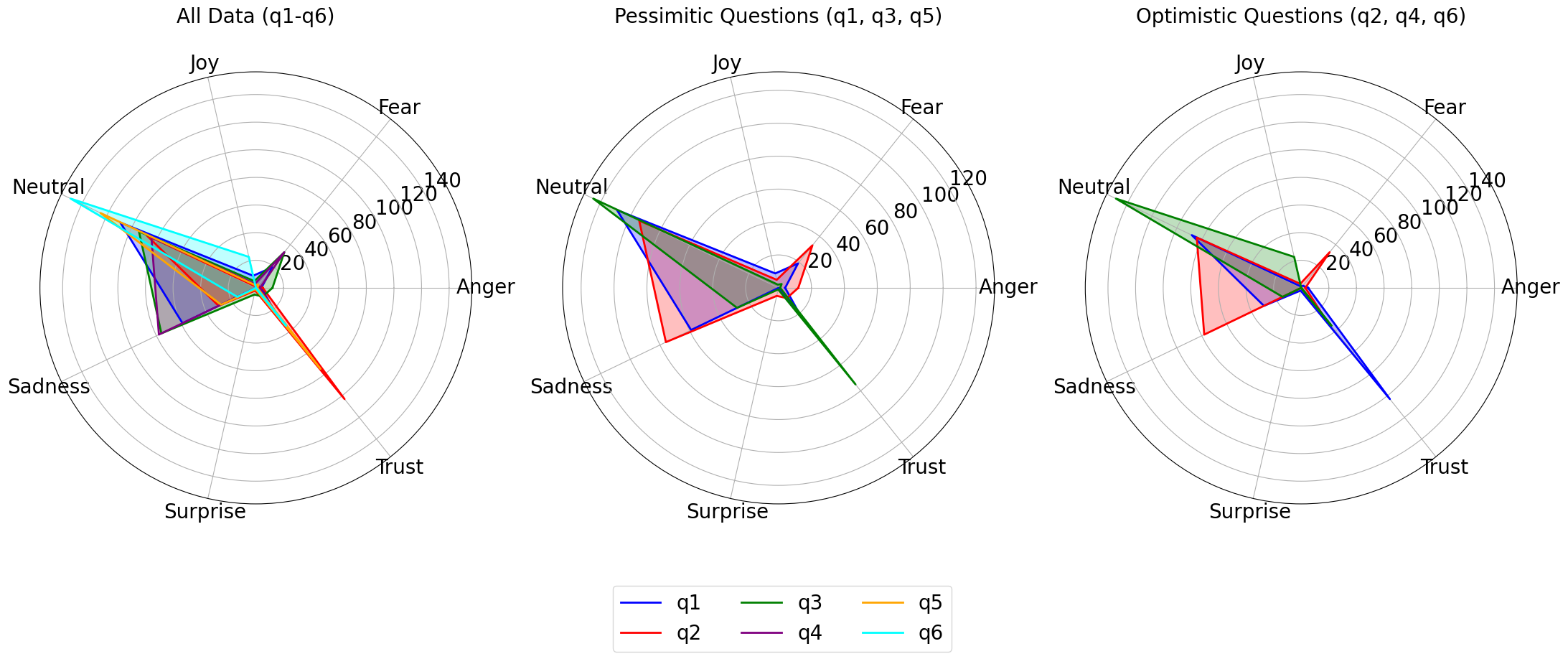}
    \caption{Emotion predictions across: all PsihoRo texts (left), negative-topic (pessimistic) questions (middle), and positive-topic (optimistic) questions (right).}
    \label{fig:all_emotions}
\end{figure*}

Using the emotion detection model described in Section \ref{sec:methodology}, we predicted emotions for all PsihoRo texts, then aggregated the results using radar charts to visualize overall emotional patterns. In Figure \ref{fig:all_emotions}, we show the emotion predictions across all PsihoRo texts, as well as for the negative-topic (pessimistic) questions (1, 3, 5) and the positive-topic (optimistic) questions (2, 4, 6). 

\begin{figure*}[h!]
    \centering
    \includegraphics[width=1\linewidth]{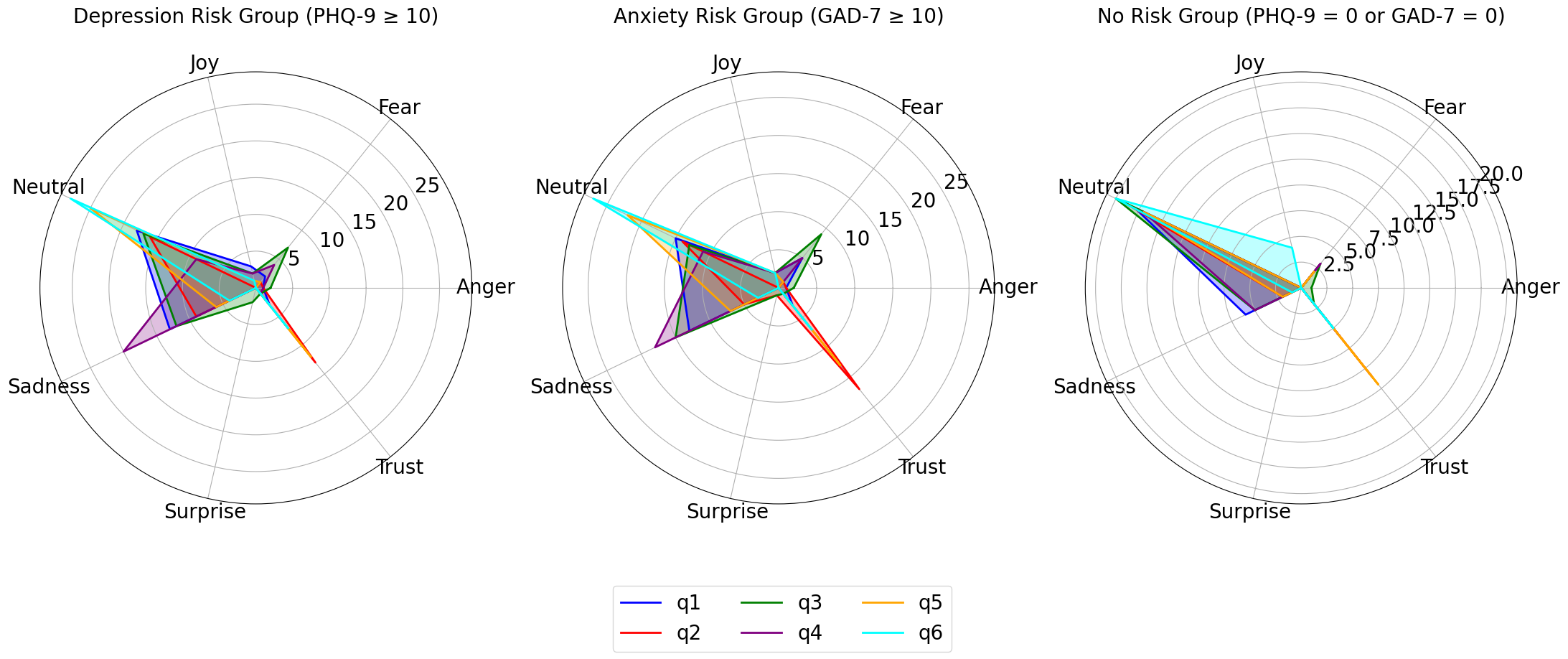}
    \caption{Emotion predictions across the three studied population groups.}
    \label{fig:emotions_in_qs}
\end{figure*}

Surprise is an emotion that is completely absent from the PsihoRo data. The following emotions are more common, in order of appearance: anger, joy, fear, sadness, trust, and, most commonly, neutral. The emotional pattern for the more pessimistic questions is displayed in the center of Figure \ref{fig:all_emotions}, while the emotional pattern for the more optimistic questions is displayed on the right. Emotions such as sadness and anger are on the rise on the Pessimistic Questions, while joy and trust are more prevalent on the Optimistic Questions.


The emotions present in the texts written by the three primary demographic groups mentioned at the beginning of Section \ref{sec:methodology} are presented in Figure \ref{fig:emotions_in_qs}. While the neutral emotion is strongly represented across all groups, the population at risk for depression shows a high prevalence of sadness. In addition, the group at risk for anxiety exhibits emotions of sadness, fear and trust.



\subsection{Topic Modeling}
We applied the topic modeling analysis described in Section \ref{sec:methodology} for each of the three population groups studied in this work. The main topics are derived from the six open questions asked in the form. However, some differences appear among the three groups of respondents, as discussed in the following subsections.

\subsubsection{No Risk Group}
\noindent\textbf{Topic 1} highlights how individuals perceive the current political and social period in a critical and sometimes negative way. The words “\textit{negative}”, “\textit{decisions}”, “\textit{politics}”, “\textit{measure}”, “\textit{news}” and “\textit{period}” indicate a concern for the effects of government decisions and how they are reflected in the media. The presence of the term “\textit{sport}” suggests a possible comparison between different areas of public life.


\noindent\textbf{Topic 2} reflects the complex interaction between information conveyed through news, political life, and how they influence the general state of the population. Terms such as “\textit{politics}”, “\textit{news}”, “\textit{state}”, “\textit{people}” and “\textit{politics}” indicate an increased attention to how media discourse shapes collective perceptions and emotions. 


\noindent\textbf{Topic 3} addresses how individuals interpret and process information related to elections and political decisions. The terms “\textit{measure}”, “\textit{depends}”, “\textit{people}”, “\textit{time}”, “\textit{elections}” and “\textit{example}” suggest a personal analysis of the electoral context and an attempt to understand the relationship between information, time and decision. 


\noindent\textbf{Topic 4} highlights the importance of authentic human interaction and the balance between the inner and social dimensions of the individual. The words ``\textit{listen}'', ``\textit{outer}'', ``\textit{inner}'', ``\textit{community}'' and ``\textit{matter}'' suggest an openness to empathetic communication and highlight the value of social belonging.


\noindent\textbf{Topic 5} focuses on the emotional dimension of everyday experiences, indicating how individuals manage their affective states and reactions in common situations. Words such as “\textit{affect}”, “\textit{try}”, “\textit{day}”, “\textit{go}” and “\textit{enough}” express a constant effort to adapt and maintain psychological balance in the face of daily pressures by self-regulation and personal reflection.


\subsubsection{Depression Risk Group}
\noindent\textbf{Topic 1} captures the perception of fatigue, daily routine and the constant effort that individuals put into their personal and professional lives. Associated terms such as “\textit{sleep}”, “\textit{hard}”, “\textit{walking}” and “\textit{people}” indicate a general state of exhaustion and the need for balance between personal and social life. 


\noindent\textbf{Topic 2} highlights the importance of psychological resilience and the ability to adapt to stressful or unpredictable situations. Words such as “\textit{resilience}”, “\textit{situation}”, “\textit{time}”, “\textit{community}” and “\textit{important}” suggest a solution-oriented approach and maintaining a state of balance in contexts of uncertainty. 


\noindent\textbf{Topic 3} focuses on how individuals relate to political processes and the electoral climate. The terms “\textit{politics}”, “\textit{elections}”, “\textit{people}”, “\textit{moment}” and “\textit{listening}” indicate both interest in political events and their influence on social attitudes and conversations. 


\noindent\textbf{Topic 4} captures the potentially harmful impact of social networks on interpersonal relationships and community cohesion. The presence of the words “\textit{negative}”, “\textit{social}”, “\textit{networks}”, “\textit{community}” and “\textit{socialization}” underlines the perception of a degradation of authentic communication and well-being caused by the digital environment. 


\noindent\textbf{Topic 5} reflects the concern for affective states and how social and economic factors contribute to the intensification of collective anxiety. The terms “\textit{feel}”, “\textit{affect}”, “\textit{cause}”, “\textit{anxiety}”, “\textit{country}” and “\textit{family}” suggest an emotional perspective on everyday experiences and the general social context. 

\subsubsection{Anxiety Risk Group}
\noindent\textbf{Topic 1} captures how major events, such as war, economic crises or socio-political changes, affect the perception of daily life. Words such as “\textit{affect}”, “\textit{crisis}”, “\textit{war}”, “\textit{elections}”, “\textit{life}” and “\textit{time}” indicate the concern for external instability and its influence on personal balance.

\noindent\textbf{Topic 2} highlights the close connection between political factors and individual emotional state. Terms such as “\textit{politics}”, “\textit{stress}”, “\textit{affect}”, “\textit{feel}”, “\textit{cause}” and “\textit{family}” suggest a direct relationship between socio-political events and the level of anxiety or psychological tension felt.

\noindent\textbf{Topic 3} addresses the role of community and collective resilience in the face of the constant flow of negative information from the media. The words “\textit{resilience}”, “\textit{news}”, “\textit{negative}”, “\textit{community}”, “\textit{important}” and “\textit{trying}” indicate an attempt to find the balance between information and protecting one’s mental state.

\noindent\textbf{Topic 4} explores the effects of social media on psychological state, with a focus on anxiety and social pressure. Terms such as “\textit{social}”, “\textit{anxiety}”, “\textit{state}”, “\textit{hard}”, “\textit{networks}” and “\textit{present}” suggest a connection between the intensive use of social networks and the degradation of emotional state. The topic reflects how virtual interactions can accentuate psychological discomfort.

\noindent\textbf{Topic 5} emphasizes the connection between people, community and the sense of belonging. The words “\textit{people}”, “\textit{community}”, “\textit{listen}”, “\textit{dear}” and “\textit{happen}” highlight the importance of close relationships and solidarity in maintaining emotional balance.

\section{Conclusion and Future Work}

In this paper, we presented PsihoRo, the first open-source dataset for mental health in Romanian.  The dataset was collected from 205 participants who filled in a comprehensive form that included both open-ended questions for text collection and validated questionnaires, specifically the PHQ-9 and GAD-7, to evaluate their risk of depression and anxiety.

We analyzed the collected data using the psychologically meaningful categories from LIWC, emotion analysis and topic modeling. We observed differences in LIWC categories related to body-related words, which may indicate somatic awareness, and tentative words, which indicate uncertainty, that correlated positively with both depression and anxiety scores. A higher preoccupation with mortality was also observed in relation to depression, while anxiety scores were positively correlated with a preoccupation with family. While LIWC analyses showed that lexical indicators can effectively capture underlying psychological patterns related to depression and anxiety, we did not find any association between the linguistic marker ``I'' and depression or anxiety in the Romanian language, consistent with previous research \cite{trifu2024linguistic}. The emotion analysis revealed a prevalence of sadness in the population group at risk for depression, as well as fear in the population group at risk for anxiety, which aligns with previous research \cite{low2020nlp, atapattu-etal-2022-emoment, Varastehnezhad2025AIIM}. Topic modeling showed that although some perceptions and emotions are captured in textual responses, a multi-perspective approach that goes beyond self-expressed text is necessary to fully comprehend these patterns. By offering high-quality data with precise labels for anxiety and depression risk, PsihoRo seeks to supplement this intricate research.

\section*{Ethical Considerations and Limitations}
The data collection received approval from the Ethics Committee of the University of Bucharest. All respondents provided their informed consent before participating in the survey, which was completely anonymous. No personal information, such as names, surnames, email addresses, or IP addresses, was requested or stored.

We believe that increasing the number of respondents through crowdsourcing or similar platforms could significantly enhance PsihoRo. We aim to implement this in future versions of this resource.

\section*{Acknowledgments}

The authors would like to thank psychologist Virginia Neacșu for her valuable feedback and suggestions throughout the development of this work. 

This work was partially supported by Druid AI, University of Bucharest and the Ministry of Education and Research, CNCS-UEFISCDI, project SIROLA, number PN-IV-P1- PCE-2023-1701, within PNCDI IV, and by the project "Romanian Hub for Artificial Intelligence - HRIA”, Smart Growth, Digitization, and Financial Instruments Program, 2021-2027, MySMIS no. 334906.

\section*{Bibliographical References}

\bibliographystyle{lrec2026-natbib}
\bibliography{lrec2026-example}



\end{document}